\documentclass[letterpaper,10pt,twocolumn,conference]{IEEEtran}

\pdfoutput=1
\usepackage[utf8]{inputenc}
\usepackage[T1]{fontenc}
\usepackage{graphicx}
\usepackage{grffile}
\usepackage{longtable}
\usepackage{wrapfig}
\usepackage{rotating}
\usepackage[normalem]{ulem}
\usepackage{amsmath}
\usepackage{textcomp}
\usepackage{amssymb}
\usepackage{capt-of}
\usepackage{hyperref}
\usepackage[margin=1in,dvips]{geometry}
\usepackage{flushend}
\usepackage{epstopdf}
\usepackage{graphicx,psfrag,amsmath,amsthm,amssymb}
\usepackage[caption=false, font=footnotesize]{subfig}
\usepackage{url,color, xcolor}
\usepackage[ruled, lined, linesnumbered]{algorithm2e}
\usepackage[hyperpageref]{backref}
\hypersetup{
colorlinks   = true, 
urlcolor     = magenta, 
linkcolor    = red, 
citecolor   = {green!50!black} 
}

\IEEEsettextwidth{0.75in}{0.75in}
\IEEEsetsidemargin{c}{0pt}

\IEEEaftertitletext{\vspace{-300\baselineskip}}

\author{\IEEEauthorblockN{Guoxiang Zhang}
\IEEEauthorblockA{\textit{School of Engineering} \\
 \textit{UC Merced}\\
 Merced, USA \\
gzhang8@ucmerced.edu}\\\vspace{-0.58in}
\and
\IEEEauthorblockN{YangQuan Chen}
\IEEEauthorblockA{\textit{School of Engineering} \\
\textit{UC Merced}\\
Merced, USA \\
ychen53@ucmerced.edu}\\\vspace{-0.58in}}

\title{\vspace{-0.06in}\Large{\textbf{More Informed Random Sample Consensus}}\vspace{0.24in}}
\hypersetup{
 pdfauthor={Guoxiang Zhang and YangQuan Chen},
 pdftitle={More Informed Random Sample Consensus},
 pdfkeywords={},
 pdfsubject={},
 pdfcreator={Org mode}, 
 pdflang={English}}
\begin{document}

\maketitle

\begin{abstract}
Random sample consensus (RANSAC) is a robust model-fitting algorithm. It is widely used in many fields including image-stitching and point cloud registration. In RANSAC, data is uniformly sampled for hypothesis generation. However, this uniform sampling strategy does not fully utilize all the information on many problems. In this paper, we propose a method that samples data with a L\'{e}vy distribution together with a data sorting algorithm. In the hypothesis sampling step of the proposed method, data is sorted with a sorting algorithm we proposed, which sorts data based on the likelihood of a data point being in the inlier set. Then, hypotheses are sampled from the sorted data with L\'{e}vy distribution. The proposed method is evaluated on both simulation and real-world public datasets. Our method shows better results compared with the uniform baseline method.
\end{abstract}


\section{Introduction}
\label{sec:orgae940ef}
The random sample consensus (RANSAC) \cite{Fischler1981} is a robust model parameter estimation algorithm that can work with data which contain a large proportion of outliers. Due to its robustness to outliers, RANSAC is widely used in many fields including signal estimation~\cite{Niedfeldt2013}, image-stitching~\cite{Choi2009}, visual odometry \cite{Kitt2010}, pattern detection~\cite{Kanazawa2004} and point cloud registration \cite{Zhang2018av,Zhang2018aw}.

However, the RANSAC algorithm requires long computation time \cite{Niedfeldt2013} because it needs to test a large number of hypotheses to find a good model. The number of hypotheses is set by a formula that depends on the number of inliers. But, in practice, the number of inliers is typically unknown. Thus RANSAC may be used with fewer iterations. Also, the probability of getting a correct model estimation dramatically decreases when the initial inlier ratio is low \cite{Li2010}. Moreover, it has been observed that a noise-contaminated outlier-free hypothesis may lead to a bad model estimate, which further requires more hypotheses to be tested \cite{Chum2003}.

In this work, we propose More Informed Random Sample Consensus (MI-RANSAC) that samples data with a L\'{e}vy distribution on ranked data. In the hypothesis sampling step of our proposed method, data is ranked with a sorting metric we proposed, which sorts data based on the likelihood of a data point being from the inlier set. Then, hypotheses are sampled from the sorted data with L\'{e}vy distribution.

This L\'{e}vy distribution-based sampling strategy improves the probability of sampling a good hypothesis thus increase the chance of finding correct solutions. On the other hand, our method can converge to correct solutions with a similar probability with a smaller number of iterations. Our experiments on simulation and real-world data confirm the advantage of our method.

\section{Related work}
\label{sec:orge8de867}

After RANSAC \cite{Fischler1981} proposed by Fischler and Bolles, research has been done to improve its performance. LO-RANSAC \cite{Chum2003}  proposes a local optimization step after the minimal sample model, which remedy an incorrect assumption that a model computed from outlier-free samples is consistent with all inliers.
In MLESAC \cite{Torr2000}, Torr and Zisserman proposed a new way of accessing model quality by choosing the solution that maximizes the likelihood rather than just the number of inliers. It is reported to be superior to the inlier counting of the plain RANSAC and less sensitive to threshold setting~\cite{Barath2018}.
A differentiable RANSAC layer is introduced in~\cite{Brachmann2017}  that can be used in neural networks in an end-to-end manner which provides promising results. Latent RANSAC \cite{Korman2018} presents an approach that can evaluate a hypothesis independent of input size. This method is based on the assumption that correct hypotheses are tightly clustered together in the latent parameter domain.

Among all the different improvements on RANSAC, a few of them are closely related to our method. PROSAC~\cite{Chum2005}  orders the set of correspondences by a similarity function. Its samples are drawn from progressively larger sets of top-ranked correspondences.
EVSAC~\cite{Fragoso2013} proposes a probabilistic parametric model that allows assigning a confidence value to each matching correspondence and thus accelerates iterations with hypothesis models.
In NAPSAC~\cite{Myatt2002} a new sampling strategy is proposed under the assumption that inliers tend to be closer to one another than outliers. 
GroupSAC~\cite{Ni2009}  assumes that there exists some grouping between features in data. The grouping can come from prior information such as optical flow based clustering. To utilize this information, a binomial mixture model is introduced for sampling. Compared with these previous works, our method has a totally different sampling strategy with L\'{e}vy distribution.

\section{Method}
\label{sec:orgdf1fd46}
The overall steps of our proposed algorithm, as described in Algorithm \ref{algo1}, has a similar structure to RANSAC. First, input data \(\mathcal{X}\) are ranked with a similarity metric. Then hypothesis samples are drawn from ranked data with a L\'{e}vy distribution rather than a uniform distribution as in the plain RANSAC. These hypotheses are tested to get a set of inlier points. Finally, the model which leads to the largest inlier set is selected as the result.  For details of the ranking and sampling steps, we describe them in \ref{ranking_sec} and \ref{sampling_sec} respectively.  
\begin{algorithm}
    \SetKwFunction{findInliers}{findInliers}
    \SetKwFunction{rank}{rank}

    \SetKwInOut{KwIn}{Input}
    \SetKwInOut{KwOut}{Output}

    \KwIn{ $\mathcal{X}$, $k_{max}, \tau$ }
    \KwOut{$\theta^*, \mathcal{I}^*$}

    $k \leftarrow 0$, $\mathcal{I}^* \leftarrow \varnothing$

    \tcp{Data ranking}    
    $\mathcal{X}^r \leftarrow \rank(\mathcal{X}, by=\text{similarityMetric})$ 

    \While{$k<k_{max}$}{
        \tcp{Hypothesis generation}
        Sample a subset of $m$ points with L\'{e}vy distribution from $\mathcal{X}^r$ \\
        Estimate model parameters $\theta_{k}$

        \tcp{Verification}

        $\mathcal{I}_k \leftarrow \findInliers(\mathcal{X}, \theta_{k}, \tau)$

        \If{$|\mathcal{I}_k| < |\mathcal{I}_{max}|$}{
            $\theta^* \leftarrow \theta_k$,  $\mathcal{I}^* \leftarrow \mathcal{I}_k$
         }
         $k \leftarrow k + 1$

    }

    \caption{MI-RANSAC algorithm}
    \label{algo1}
\end{algorithm}

\subsection{Data ranking}
\label{sec:org01869c2}
\label{ranking_sec}
In many RANSAC use cases, input data include matched point correspondence, such as in image matching and point cloud registration. For these cases, there is information useful to exploit. In our MI-RANSAC, we rank these correspondence pairs in descending order in terms of likelihood to be an inlier pair. Ranking data helps to put good pairs of point matches to the front and the ones that are more likely to be the wrong ones to the end. Different ranking methods can be used in our MI-RANSAC. Here, we use a simple yet efficient ranking metric.  Given a pair of feature vectors \((x_1, x_2)\) from two points. We have
\begin{equation}
r = |\left\Vert(x_1 - x_2)\right \Vert^2 < \tau_d|.
\label{eq:ranking}
\end{equation}
Equation \eqref{eq:ranking} maps the feature vector pair to a metric space \(r \in R\). such that \(r>r^{\prime}\) when \(p(x_1, x_2) > p(x^{\prime}_1, x^{\prime}_2)\), where \(p(x_1, x_2)\) denotes the probability of the point correspondence pair being a correct one. 

\begin{figure}[h]\centering
\begin{center}
\includegraphics[width=0.8\linewidth]{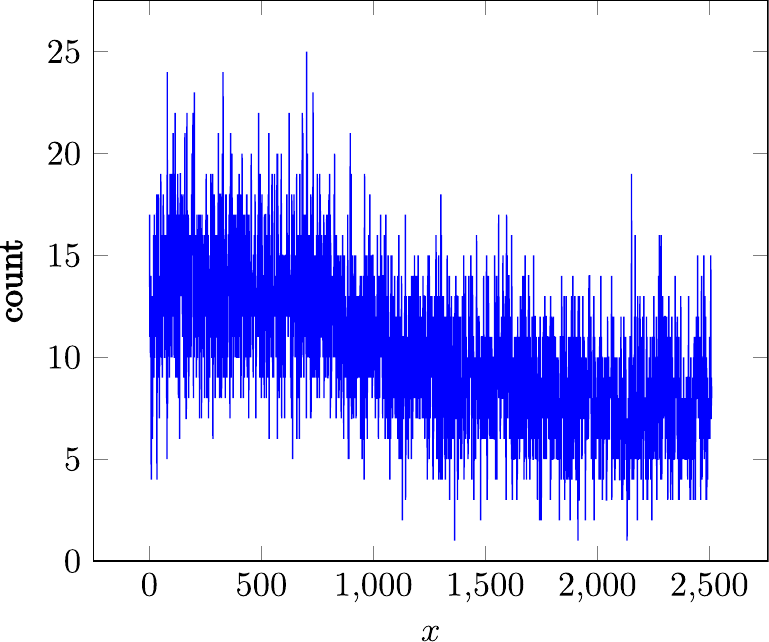}
\end{center}
\caption{The correctness histgram on different indices. Results are from the livingroom 2 sequence of the Augmented ICL-NUIM dataset~\cite{Choi2015}.}\label{fig_freq_last}
\end{figure}

We run experiments on Augmented ICL-NUIM dataset~\cite{Choi2015} and generate groundtruth correspondences by locating nearest neighbors after transforming point clouds to the world frame. We show the ranking results of this handcrafted ranking metric by visualizing the correct ratio at a specific index. We get Fig. \ref{fig_freq_last}. 

\subsection{L\'{e}vy sampling}
\label{sec:orgebf0b58}
\label{sampling_sec}

L\'{e}vy Distribution is a probability distribution that is both continuous, for non-negative random variables, and stable. It follows probability distribution function as in \eqref{eq_levy}.
\begin{equation}
\label{eq_levy}
f(x;\mu,c)=\sqrt{\frac{c}{2\pi}}~~\frac{e^{ -\frac{c}{2(x-\mu)}}} {(x-\mu)^{3/2}},
\end{equation}
where \(c\) is the scale parameter (\(c > 0\)) and \(\mu\) is the location parameter. It is meaningful when \(x \ge \mu\).

We sample from L\'{e}vy distribution such that it can give more weight on top-ranked correspondences while not leave the less likely ones behind. It can also approximate uniform distribution so that it can degenerate to the plain RANSAC in the worst case. 

\begin{figure}[htbp]
\centering
\includegraphics[width=0.8\linewidth]{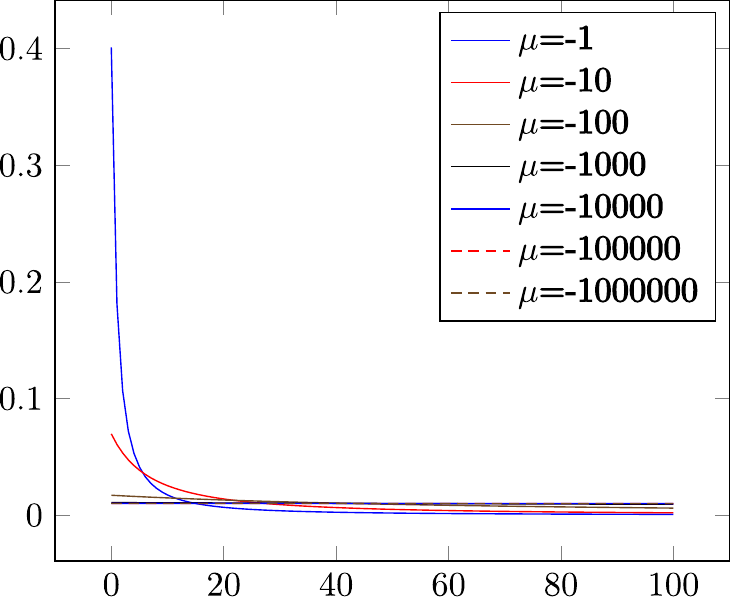}
\caption{PDF of L\'{e}vy distribution within  \(\left(0, 100\right)\) for \(c=1\) with different parameter \(\mu\) \label{fig_fc_pdf}}
\end{figure}

\begin{figure}[htbp]
\centering
\includegraphics[width=0.8\linewidth]{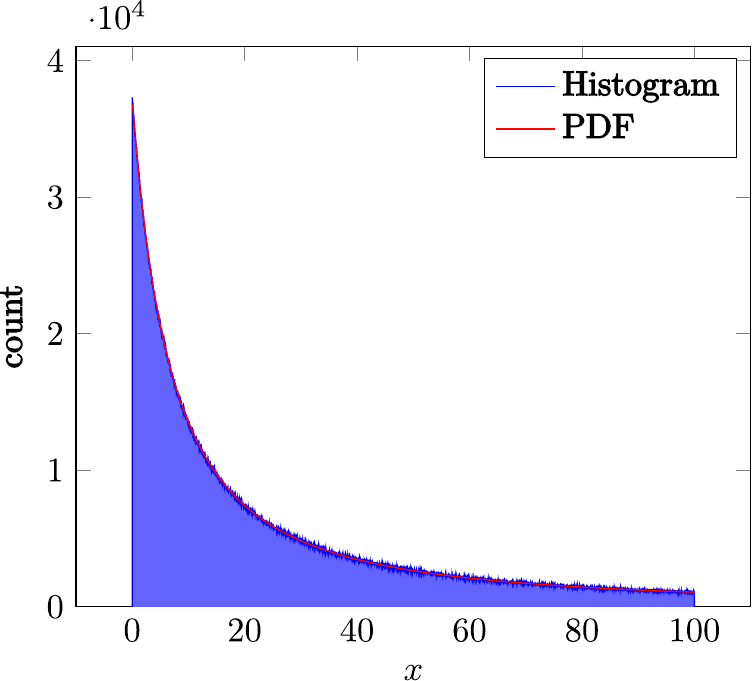}
\caption{Histogram of interger indices sampled from L\'{e}vy distribution within  \(\left(0, 100\right)\) for \(c=1\), \(\mu = -10\) \label{fig_fc_pdf_hist}}
\end{figure}

A special step needs to be considered because hypothesis sampling is integer index sampling. These indices are within a range. To solve this, we sample from truncated L\'{e}vy distribution in range \([0, m]\). Then we scale the random numbers to \([0, n]\) and round them to the nearest integer. With the truncated L\'{e}vy distribution that follows a probability distribution function (PDF) as \eqref{eq_levy}, we can get PDFs of different shape if they have different parameters. In Fig. \ref{fig_fc_pdf}, when we truncate and normalize the PDFs within \(\left(0, 100\right)\) with different \(\mu\), we get different curves. When \(\mu \rightarrow \infty\), the curve is flatter thus more similar to the PDF of a uniform distribution.  
We also plot a histogram of sampled indices using the described sampling method. The plots are in Fig. \ref{fig_fc_pdf_hist}. In this experiment, \(400000\) hypotheses with each hypothesis containing \(4\) point correspondences are sampled.

\section{Experiments}
\label{sec:org0cae615}

In order to evaluate our method, experiments are performed on both real and synthetic data.
\subsection{Simulation}
\label{sec:orgdec2d34}
\label{sec_sim}

\begin{figure}[!htbp]
\centering
\includegraphics[width=0.95\linewidth]{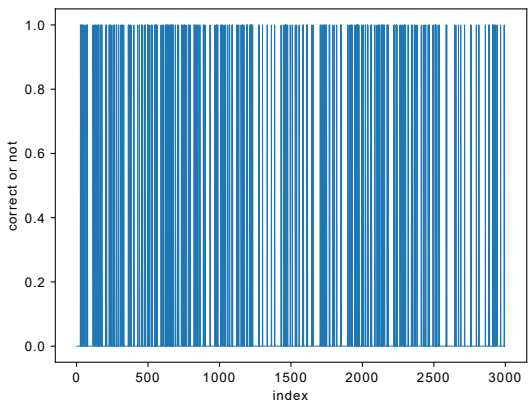}
\caption{Simulation data visualization}
\label{fig:sim_data_vis}
\end{figure}

In order to test the performance of different hypotheses sampling strategies, we create a simulation that is informative and easy to perform. In this simulation, a vector of boolean is generated as input data. This vector simulates a set of feature correspondences with \texttt{true} as an inlier pair and \texttt{false} as an outlier pair. This vector is generated in a way that the indices of \texttt{true} elements follow truncated L\'{e}vy distribution. A visualization of one instance of such data is in Fig. \ref{fig:sim_data_vis}, in which \texttt{true} elements are visualized as vertical lines. The data is generated such that vertical lines are denser on the left and sparser on the right to simulate ranked data.

In our experiments, we create boolean input vector data using L\'{e}vy distribution with \(\mu=-10\) and \(c=1\). The boolean vector will have \texttt{true} value on L\'{e}vy sampled index positions and \texttt{false} on all the other locations. We use \(3000\) as the length of data and \(300\) is the number of inliers.

We evaluate eight different hypothesis sampling strategies:
\begin{itemize}
\item Uniform distribution
\item L\'{e}vy distribution with \(c = 1\), \(\mu = \{-1, -50, -10^2, -10^3, -10^4, -10^5, -10^6 \}\)
\end{itemize}
We also change number of samples in a hypothesis from  \(\{2, 3, 4\}\) and number of hypotheses from \(\{10^2, 10^3, 10^4, 10^5, 10^6\}\) to explore performance difference under different coverage settings.

\begin{figure*}[htb]\centering
\subfloat[Sample size 2]{\includegraphics[width=0.32\textwidth]{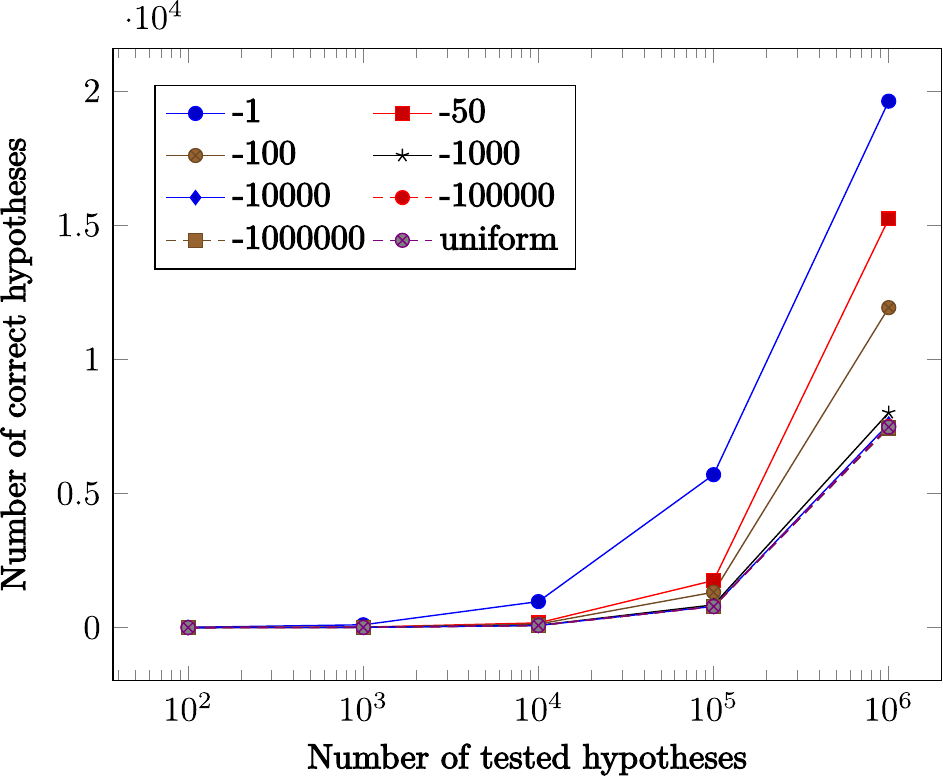}}
\subfloat[Sample size 3]{\includegraphics[width=0.31\textwidth]{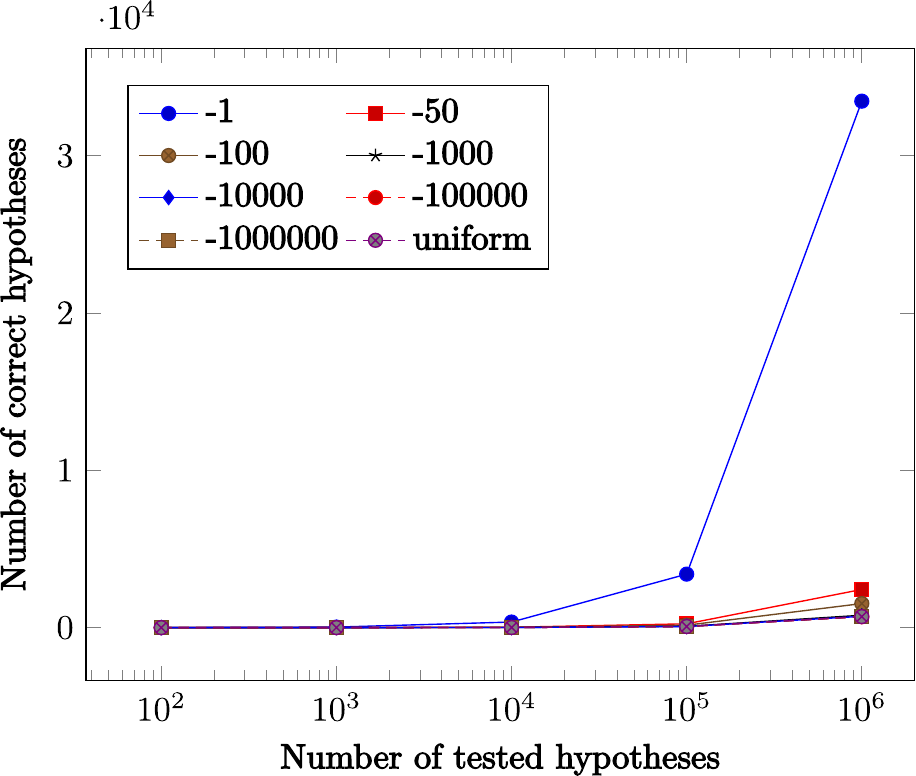}}
\subfloat[Sample size 4]{\includegraphics[width=0.32\textwidth]{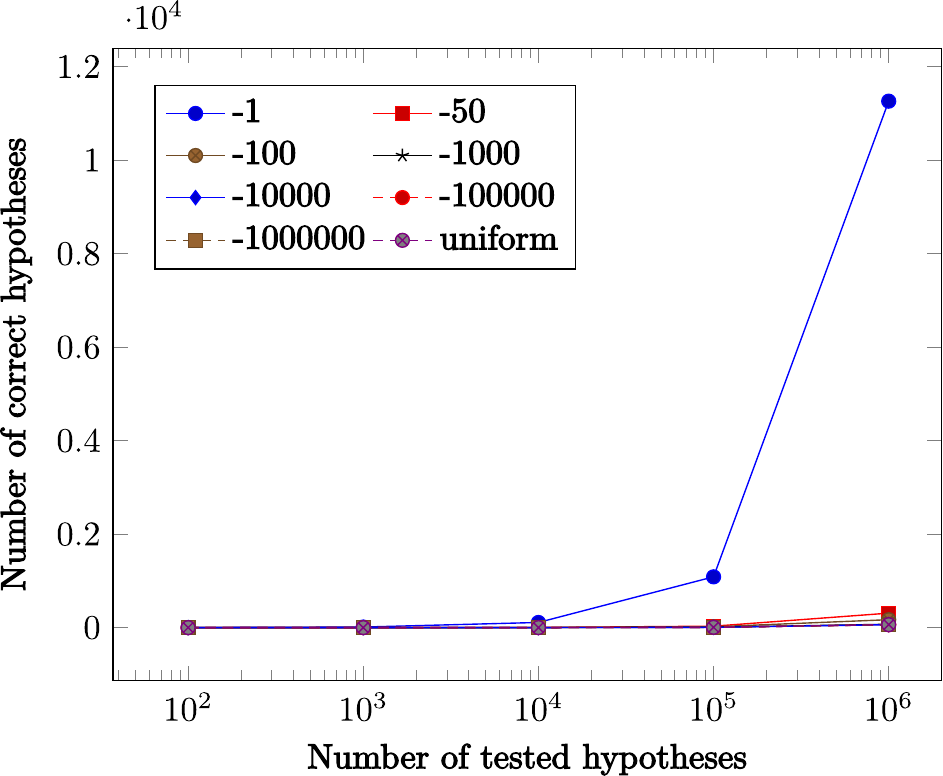}}
\caption{Number of unique outlier-free hypotheses under difference experiment configurations}\label{fig:sim_res}
\end{figure*}

We run experiments with all the configurations specified above and plot results in Fig. \ref{fig:sim_res}. Under all the different sampling strategies, L\'{e}vy distributions with \(\mu=-1\) performs the best, much better than uniform sampling distribution. In Fig. \ref{fig:sim_res} (a), when number of hypotheses go up, the results become less diverge. But when the hypotheses number is low, a more skewed distribution leads to a better result. More interestingly, In Fig. \ref{fig:sim_res} (c), the performance is better with \(10^5\) hypotheses tested when \(u=-1\) than the rest of \(u\) values even with an order of magnitude more hypotheses tested. 

\subsection{Point cloud registration}
\label{sec:orgfce9ca7}

We further evaluate our method on the point cloud registration problem. We use the augmented ICL-NUIM point cloud registration dataset \cite{Choi2015} to quantitatively analysis performance of our method. This dataset has four sets of point clouds and groundtruth rigid transformation between all possible point cloud pairs that overlap more than 30\%. The evaluation metrics are precision and recall defined as in \eqref{eq:pr}:
\begin{equation}
\label{eq:pr}
\begin{split}
\text{precision} & =\frac{TP}{TP+FP}\\
\text{recall} & =\frac{TP}{TP+FN}
\end{split}
\end{equation}
where \(TP\) is true positive; \(TN\) is true negative; \(FP\) is false positive and \(FN\) is false negative. Both precision and recall are the higher the better.

\begin{figure}[!hbt]\centering
\subfloat[Recall]{\includegraphics[width=0.8\linewidth]{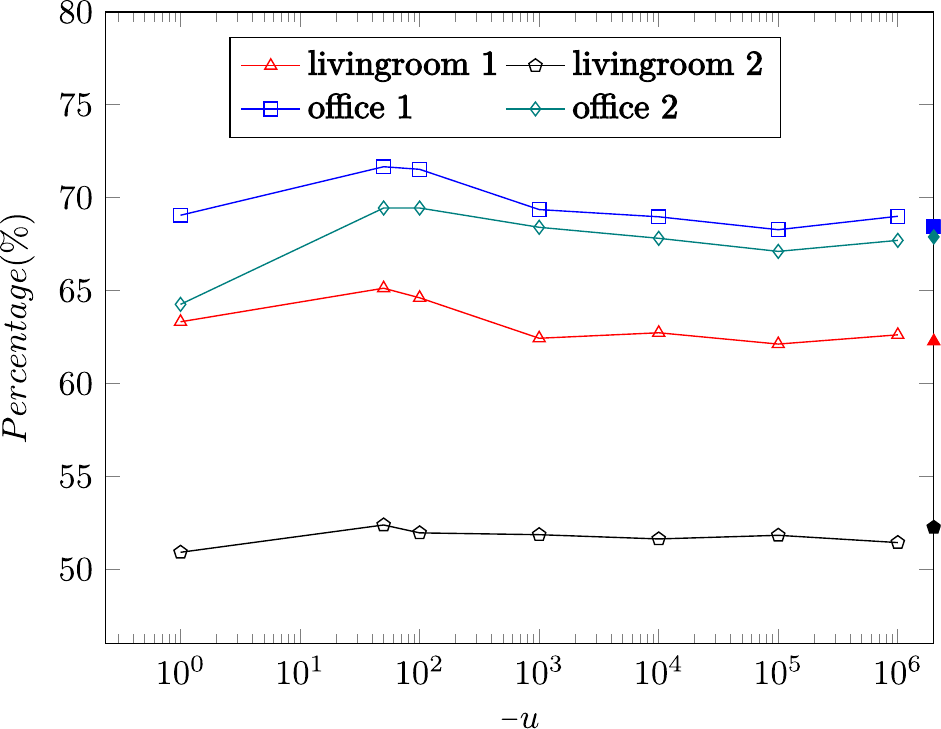}}\\
\subfloat[Precision]{\includegraphics[width=0.8\linewidth]{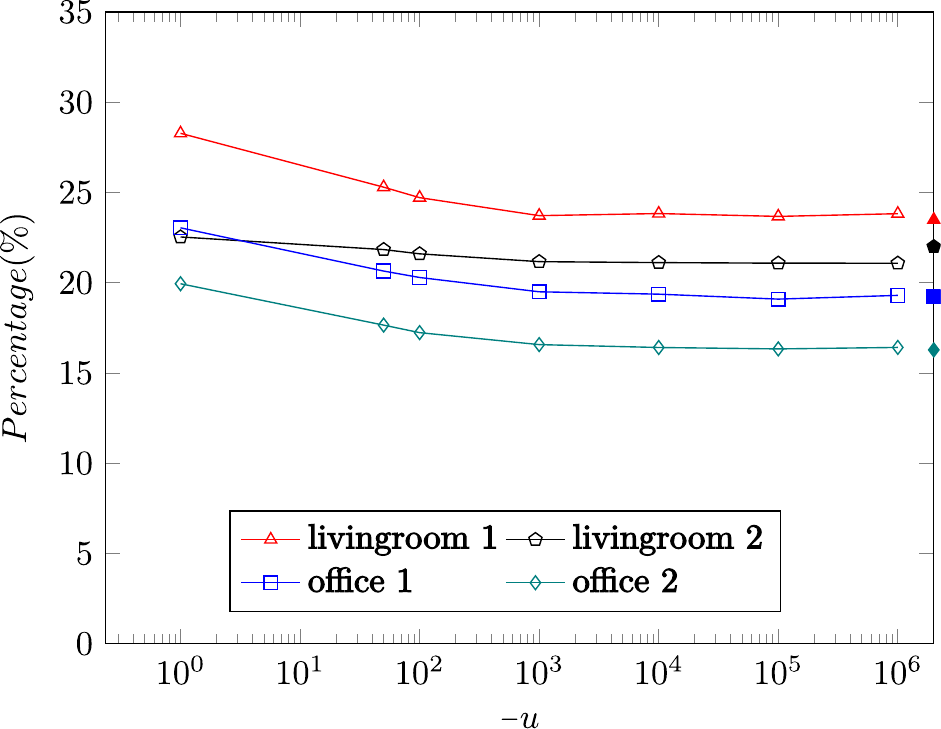}}
\caption{Recall and precision of RANSAC on Aug-ICL-NUIM data with our handcrafted ranking method. All the performance results are from the average of $20$ Monte Carlo runs. For each subfigure, the marks on the right most axis are the performance of the uniform distribution.}\label{fig_pr_vs_mu}
\end{figure}

We compare the performance of our MI-RANSAC system to the baseline RANSAC that samples from a uniform distribution. We run experiments on this dataset with different sampling distributions. We use the same setting as in section \ref{sec_sim}: eight L\'{e}vy distributions and one uniform distribution. Experiments are repeated for \(20\) times to remove the influence of randomness on the final results. Results are shown in Fig. \ref{fig_pr_vs_mu}. Each curve is for a point cloud set.

In this result, we value more on a better recall because precision can be improved by post-processing but recall cannot.  We can see that there are a few \(\mu\) values that lead to better recall results than the baseline. There is consistency in the performance along \(\mu\). The performance starts low when \(\mu\) is very small. Then it goes up when \(\mu\) is around \(-50\). Then converge to the performance of uniform distribution. From this figure, we can see a clear performance benefit. More importantly, from the performance curves, we notice that the best value for \(\mu\) is consistent across all the data sequences, which means that the method generalizes well on different data.

\section{Conclusion and future work}
\label{sec:org4ffff9f}

In this work, we propose MI-RANSAC that samples data with a L\'{e}vy distribution after data ranking. The proposed method is evaluated on both simulation and real-world public datasets. In experiments, our method shows better results compared with the uniform baseline method.

In the future, we will explore more on the effect of changing the scale parameter \(c\) in the L\'{e}vy distribution. Another effort is to explore potential performance gain on real-world LiDAR-based point cloud data processing.

\bibliographystyle{IEEEtran}
\bibliography{compact_lib.bib}

\end{document}